\documentclass{article}

\usepackage[final,nonatbib]{neurips_2022}

\usepackage[utf8]{inputenc} 
\usepackage[T1]{fontenc}    
\usepackage{hyperref}       
\usepackage{url}            
\usepackage{booktabs}       
\usepackage{amsfonts}       
\usepackage{nicefrac}       
\usepackage{microtype}      
\usepackage{xcolor}         

\usepackage{amsmath}
\usepackage{amssymb}
\usepackage{mathtools}
\usepackage{amsthm}
\usepackage{dsfont}
\usepackage{array,multirow}
\usepackage[para,online,flushleft]{threeparttable}

\usepackage{algorithm}
\usepackage{algorithmic}

\theoremstyle{plain}
\newtheorem{theorem}{Theorem}[section]

\theoremstyle{definition}
\newtheorem{definition}[theorem]{Definition}

\theoremstyle{remark}

\usepackage{color}
\usepackage{xspace}
\usepackage{url}
\usepackage{pifont} 

\usepackage{silence} 
\usepackage[normalem]{ulem} 

\usepackage{enumitem} 
\setlist{nosep} 

\usepackage{wrapfig}
\usepackage{rotating} 

\usepackage{booktabs} 

\usepackage{array} 
\newcommand{\PreserveBackslash}[1]{\let\temp=\\#1\let\\=\temp}
\newcolumntype{C}[1]{>{\PreserveBackslash\centering}p{#1}}
\newcolumntype{R}[1]{>{\PreserveBackslash\raggedleft}p{#1}}
\newcolumntype{L}[1]{>{\PreserveBackslash\raggedright}p{#1}}

 \usepackage{titlesec}
 \titlespacing\section{0pt}{0pt plus 0pt minus 2pt}{0pt plus 0pt minus 2pt}
 \titlespacing\subsection{0pt}{0pt plus 0pt minus 2pt}{0pt plus 0pt minus 2pt}
 \titlespacing\subsubsection{0pt}{0pt plus 0pt minus 2pt}{0pt plus 0pt minus 2pt}

\newcommand{\syslong}[0]{Certified Robustness Transfer\xspace}
\newcommand{\sys}[0]{CRT\xspace}

\newcommand{\cifar}[0]{CIFAR-10\xspace}
\newcommand{\imagenet}[0]{ImageNet\xspace}

\newcommand{\smoothadv}[0]{SmoothAdv\xspace}
\newcommand{\macer}[0]{MACER\xspace}
\newcommand{\smoothmix}[0]{SmoothMix\xspace}

\newcommand{\iresnets}[0]{ResNet18\xspace}
\newcommand{\iresnetl}[0]{ResNet50\xspace}
\newcommand{\cresnets}[0]{ResNet20\xspace}
\newcommand{\cresnetl}[0]{ResNet110\xspace}
\newcommand{\resnext}[0]{ResNeXt29-2x64d\xspace}
\newcommand{\dla}[0]{DLA\xspace}
\newcommand{\regnet}[0]{RegNetX\_200MF\xspace}
\newcommand{\vit}[0]{ViT\xspace}
\newcommand{\iregnet}[0]{RegNetX-4.0G\xspace}

\newif\ifsubmit
\submitfalse

\ifsubmit
    \newcommand{\amir}[1]{}
    \newcommand{\kevin}[1]{}
    \newcommand{\pratik}[1]{}

\else
    
    \newcommand{\amir}[1]{\textcolor{blue}{Amir: #1}}
    \newcommand{\kevin}[1]{\textcolor{green}{Kevin: #1}}
    \newcommand{\pratik}[1]{\textcolor{magenta}{Pratik: #1}}

\fi

\newcommand{\paratitle}[1]{\noindent\textbf{\textit{#1.}}\xspace}

\newcommand{\ie}[0]{\emph{i.e.,}\xspace}
\newcommand{\etal}[0]{\emph{et~al.}\xspace}
\newcommand{\eg}[0]{\emph{e.g.,}\xspace}

\usepackage{cleveref} 

\crefformat{section}{\S#2#1#3} 
\crefformat{subsection}{\S#2#1#3}
\crefformat{subsubsection}{\S#2#1#3}

\title{Accelerating Certified Robustness Training via Knowledge Transfer}

\author{%
   Pratik Vaishnavi\\
   Stony Brook University\\
   \texttt{pvaishnavi@cs.stonybrook.edu} \\
  \And
  Kevin Eykholt \\
  IBM Research \\
  \texttt{kheykholt@ibm.com} \\
  \AND
  Amir Rahmati\\
  Stony Brook University\\
  \texttt{amir@cs.stonybrook.edu} \\
}

\begin{document}

\maketitle

\begin{abstract}
Training deep neural network classifiers that are \textit{certifiably} robust against adversarial attacks is critical to ensuring the security and reliability of AI-controlled systems. Although numerous state-of-the-art certified training methods have been developed, they are computationally expensive and scale poorly with respect to both dataset and network complexity. Widespread usage of certified training is further hindered by the fact that periodic retraining is necessary to incorporate new data and network improvements. In this paper, we propose \syslong (\sys), a general-purpose framework for reducing the computational overhead of any certifiably robust training method through knowledge transfer. Given a robust teacher, our framework uses a novel training loss to transfer the teacher's robustness to the student. We provide theoretical and empirical validation of \sys. Our experiments on \cifar show that \sys speeds up certified robustness training by $8 \times$ on average across three different architecture generations while achieving comparable robustness to state-of-the-art methods. We also show that \sys can scale to large-scale datasets like \imagenet.
\end{abstract}

\section{Introduction}
\label{sec:intro}
\begin{figure}[t]
\begin{center}
\centerline{
    \includegraphics[width=.85\columnwidth]{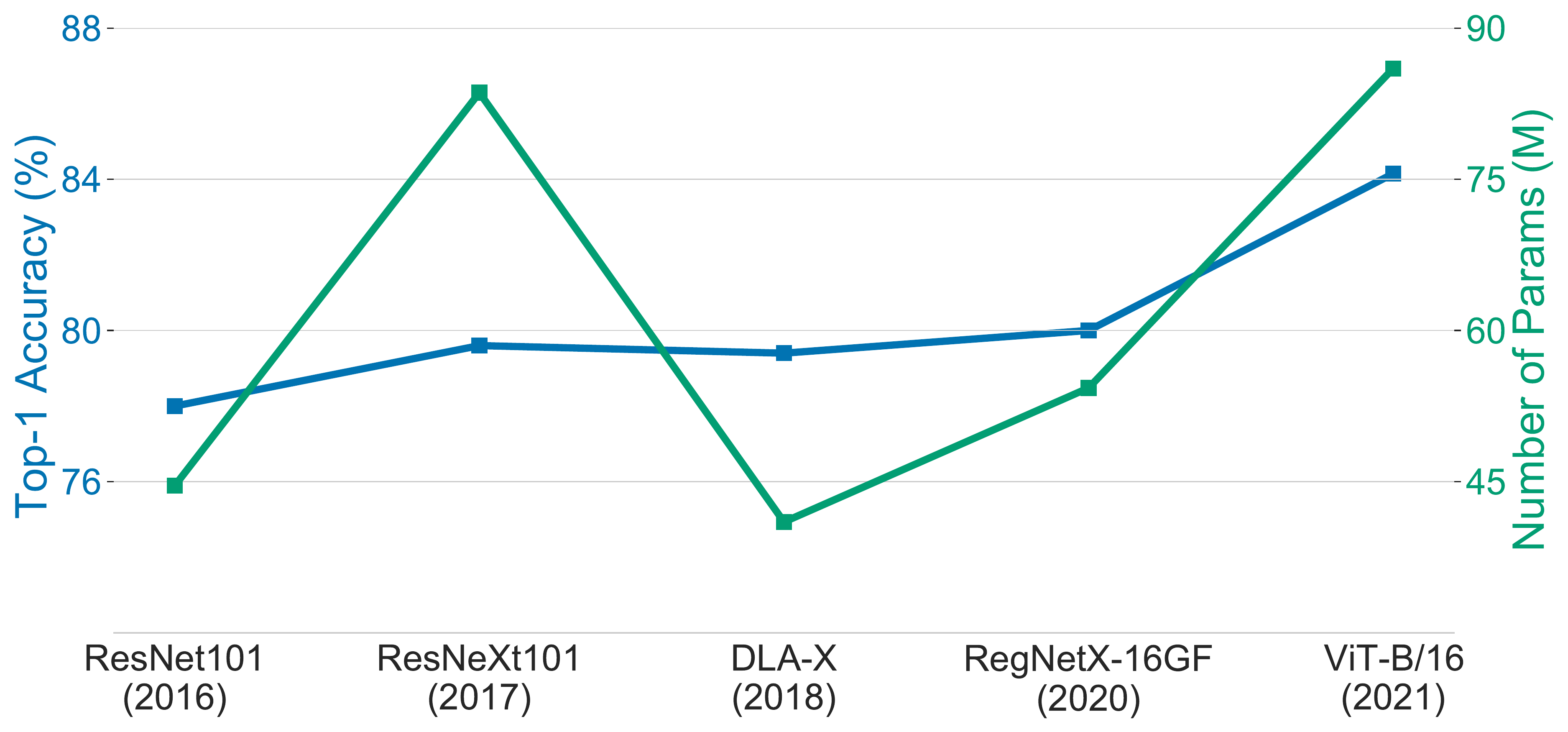}
}
\caption{Evolution of DNN architectures on the \imagenet dataset. We plot the performance (top-1 accuracy) and the number of parameters of a few popular architectures (year of release is noted in brackets). Newer generations attempt to improve performance and/or reduce network parameters.}
\label{fig:model-evolution}
\end{center}
\vspace{-2.5em}
\end{figure}

Deep Neural Networks (DNNs) are susceptible to adversarial evasion attacks~\cite{szegedy2014intriguing,goodfellow2014explaining}, that add a small amount of carefully crafted imperceptible noise to an input to reliably trigger misclassification. As a defense, numerous training methods have been proposed~\cite{madry2018towards,zhang2019theoretically,xie2019feature} to grant a classifier empirical robustness. But in the absence of any provable guarantees for this robustness, these defenses were frequently broken~\cite{athalye2018obfuscated,tramer2020adaptive}. These failures have motivated the development of training methods that grant certifiable/provable robustness to a classifier, hence safeguarding them against all attacks (known or unknown) within a pre-determined threat model. Such methods are broadly categorized as either \textit{deterministic} or \textit{probabilistic}~\cite{li2020sok}. Deterministic robustness training methods~\cite{hein2017formal,mirman2018differentiable,wong2018provable,wong2018scaling,raghunathan2018certified,gowal2019scalable,zhang2019towards,singla2020second} rely on computing provable bounds on the output neurons of a classifier for a given perturbation budget in the input space. However, the  deterministic robustness guarantees provided by these methods come at a high computational cost. Probabilistic robustness training methods address this limitation by providing highly probable (\eg with 0.99 probability) robustness guarantees at a greatly reduced computational cost. Within this category, \textit{randomized smoothing}-based methods~\cite{lecuyer2019certified,cohen2019certified, salman2019provably,li2019certified,lee2019tight,dvijotham2019framework,yang2020randomized,zhai2020macer,jeong2020consistency,jeong2021smoothmix} are considered the state-of-the-art for certifiable robustness in the $\ell_2$-space. Even so, these training methods remain an order of magnitude slower than standard training. In commercial applications where constant model re-deployment occurs to provide improvements (see Figure~\ref{fig:model-evolution}), re-training using computationally expensive methods is burdensome.

In this work, we reduce the training overhead of randomized smoothing-based robustness training methods with minimal impact on the robustness achieved. We propose \syslong (\sys), a knowledge transfer framework that significantly speeds up the process of training $\ell_2$ certifiably robust image classifiers. Given a pre-trained classifier that is certifiably robust (\ie \textit{teacher}), \sys trains a new classifier (\ie \textit{student}) that has comparable levels of robustness in a fraction of the time required by state-of-the-art methods. \sys brings down the cost of training certifiably robust image classifiers to be comparable to standard training while preserving state-of-the-art robustness.  On \cifar, \sys speeds up training by an average of $\mathbf{8}\times$ across three different architecture generations compared to a state-of-the-art robustness training method~\cite{jeong2021smoothmix}. Furthermore, we show that state-of-the-art robustness training is only necessary to train the initial classifier. Afterward, \sys can be continuously reused to transfer robustness in order to expedite future model re-deployments and greatly reduce costs associated with computational resources. Our \textbf{contributions} can be summarized as follows:
\begin{itemize}
    \setlength\itemsep{1em}
    \item We present \syslong (\sys), the first framework, to our knowledge, that can transfer the robustness of a certifiably robust teacher classifier to a new student classifier. \sys greatly reduces the time required to train certifiably robust image classifiers relative to existing state-of-the-art methods while achieving comparable or better robustness.
    \item We provide a theoretical understanding of \sys, showing how our approach of matching outputs enables robustness transfer between the student and teacher irrespective of the certified robustness training method used to train the teacher.
    \item On \cifar, we show that \sys trains certifiably robust classifiers on average $8 \times$ faster than a state-of-the-art method while having comparable or better Average Certified Radius (by $8\%$ in the best case). Furthermore, \sys reduces the cumulative computational cost of training three classifiers by $87.84\%$.
    \item We also show that \sys can be reused in a recursive manner, thus supporting a continuous re-deployment scenario (\eg in commercial applications). Finally, we show that \sys remains effective on a large-scale dataset, \imagenet.
\end{itemize}

\section{Background} \label{sec:background}
In this section, we briefly introduce certified robustness and discuss notable existing methods for training certifiably robust image classifiers using randomized smoothing.

\subsection{Preliminaries}
\paratitle{Problem Setup}
Consider a neural network classifier $f$ parameterized by $\theta$ (denoted $f_\theta$) trained to map a given input $x \in \mathds{R}^d$ to a set of discrete labels $\mathcal{Y}$ using a set of \textit{i.i.d.} samples ${\mathcal{S}=\{(x_1,y_1), (x_1,y_1), \cdots, (x_n,y_n)\}}$ drawn from a data distribution $\mathcal{D}$. The output of the classifier can be written as ${f_\theta(x)=\arg\max_{c\in\mathcal{Y}}\; z^c_\theta(x)}$. Here $z_\theta(x)$ is the softmax output of the classifier and $z^c_\theta(x)$ denotes the probability that image $x$ belongs to class $c$.

\paratitle{Certified Robustness via Randomized Smoothing}
The robustness of the classifier $f_\theta$ for a given input pair $(x,y)$ is defined using the radius of the largest $\ell_2$ ball centered at $x$ within which $f_\theta$ has a constant output $y$. This radius is referred to as \textit{robust radius} and it can mathematically be expressed as:
\begin{align} \label{eq:rob_rad}
    R(f_\theta;x,y) = \begin{cases}
    \underset{f_\theta(x') \neq f_\theta(x)}{\inf} \| x' - x \|_2 &\text{, when $f_\theta(x)=y$}\\
    \hfil 0 &\text{, when $f_\theta(x) \neq y$}
    \end{cases}
\end{align}

Within this $\ell_2$-neighborhood of $x$, $f_\theta$ is considered to be \textit{certifiably robust}. Therefore, to improve the robustness of a classifier, one needs to maximize this robust radius corresponding to any point sampled from the given data distribution. Directly maximizing the robust radius of a DNN classifier is an NP-hard problem~\cite{katz2017reluplex}. Therefore, several prior works attempt to derive a lower bound for the robust radius~\cite{li2018second,lecuyer2019certified,cohen2019certified}. This lower bound, often termed as the \textit{certified radius}, satisfies the following condition: $0\leq CR(f_\theta;x,y)\leq R(f_\theta;x,y)$, for any $f_\theta$, ($x$, $y$). In this paper, we utilize the certified robustness framework derived by Cohen~\etal~\cite{cohen2019certified} using randomized smoothing. Given a classifier $f_{\theta}$, they first define the smooth classifier $g_{\theta}$ as:

\begin{definition} \label{def:smooth}
For a given (base) classifier $f_{\theta}$ and $\sigma>0$, the smooth classifier $g_{\theta}$ corresponding to $f_{\theta}$ is defined as follows:
\begin{align} \label{eq:smooth_clf}
    g_{\theta}(x) = \underset{c\in\mathcal{Y}}{\arg\max}\, P_{\eta\sim\mathcal{N}(0,\sigma^2I)}(f_\theta(x + \eta) = c)
\end{align}
\end{definition}

Simply put, $g_{\theta}$ returns the class $c$, which has the highest probability mass under the Gaussian distribution $\mathcal{N}(x, \sigma^2I)$. Using Theorem~\ref{thm:rs}, they proved that if the smooth classifier correctly classifies a given input $x$, it is certifiably robust at $x$. They also provided an analytical form of the $\ell_2$ certified radius at $x$.


\begin{theorem} \label{thm:rs}
Let $f_{\theta}:\mathds{R}^d \mapsto \mathcal{Y}$ be a classifier and $g_{\theta}$ be its smoothed version (as defined in Definition~\ref{def:smooth}). For a given input $x\in\mathds{R}^d$ and corresponding ground truth $y\in\mathcal{Y}$, if $g_{\theta}$ correctly classifies $x$ as $y$, \ie

\vspace{-1em}
\begin{equation} \label{eq:rs_cond}
    P_\eta(f_{\theta}(x + \eta) = y) \geq \underset{y' \neq y}{\max}\, P_\eta(f_{\theta}(x + \eta) = y')
\end{equation}

then $g_{\theta}$ is provably robust at $x$ within the certified radius $R$ given by:
\vspace{-.5em}
\begin{align} \label{eq:certified_radius}
\begin{split}
    CR(g_{\theta}; x, y) = \frac{\sigma}{2} [\Phi^{-1}(P_\eta(f_{\theta}(x + \eta) = y)) - \Phi^{-1}(\underset{y' \neq y}{\max}\, P_\eta(f_{\theta}(x + \eta) = y'))]
\end{split}
\end{align}
where $\Phi$ is the c.d.f. of the standard Gaussian distribution.
\end{theorem}
This certified radius is a \textit{tight} lower bound of the robust radius defined in Equation~\ref{eq:rob_rad}, \ie it is impossible to certify $g_\theta$ at $x$ for a radius larger than $CR$.

\subsection{Training Methods for Maximizing Certified Radius}
\label{sec:cert_training_method}

In addition to the theoretical framework discussed above, Cohen~\etal~\cite{cohen2019certified} also propose a simple yet effective method for training the base classifier in a way that maximizes the $\ell_2$ certified radius of the smooth classifier, as expressed in Equation~\ref{eq:certified_radius}. We include an evaluation of their method in Appendix~\ref{app:gda_results}.
Following their work, several other works build upon the randomized smoothing framework and propose training methods that better maximize the $\ell_2$ certified radius of the smooth classifier. Salman~\etal~\cite{salman2019provably} proposed combining adversarial training ~\cite{madry2018towards} with randomized smoothing (called \smoothadv). They adapted the vanilla PGD attack to target the smooth classifier $g_{\theta}$ instead of the base classifier $f_{\theta}$. Zhai~\etal~\cite{zhai2020macer} proposed a new robustness loss, a hinge loss that enforces maximization of the soft approximation of the certified radius. Their method (called \macer) is faster than \smoothadv as it does not use adversarial training. More recently, Jeong~\etal~\cite{jeong2021smoothmix} proposed training with a convex combination of samples along the direction of adversarial perturbation for each input to regularize over-confident predictions. Their method (called \smoothmix) is the current state-of-the-art in the domain of $\ell_2$ certified robust image classifiers. Finally, we note the Consistency regularization method proposed by Jeong~\etal~\cite{jeong2020consistency}, which adds a regularization loss to existing methods that helps better maximize the certified radius.

\newpage

\section{Maximizing Certified Radius via Knowledge Transfer} 
\label{sec:method}
Although prior works have proposed methods for increasing the certified radius of the smooth classifier, their training overhead is significant, making them much slower than standard training. As we show in Table~\ref{tab:std-training-comp}, training a certifiably robust \cresnetl classifier to convergence using \smoothadv, \macer, and \smoothmix is $46.20\times$, $20.86\times$, and $4.97\times$ slower, respectively, compared to training a non-robust classifier with standard training.

Given constant innovations in architecture design (Figure~\ref{fig:model-evolution}) and the influx of new data, which may result in various tweaks to deployed networks that elicit retraining, the large overhead of state-of-the-art robustness training methods makes preserving certified robustness across model re-deployment difficult. Therefore, we propose \syslong (\sys), a training method that improves the usability of certified robustness training methods by dramatically reducing their training overhead while preserving the certified robustness. Given the \textit{base classifier} of a pre-trained certifiably robust \textit{smooth classifier}, we leverage the knowledge transfer framework to guide the training of a new base classifier (and its associated robust smooth classifier).\footnote{If no pre-trained classifier is available, we first train an architecture of lower complexity (\ie fast to train) compared to the target architecture (Section \ref{sec:teacher_unavailable}).} In this section, we describe our method and provide theoretical justification for its effectiveness.



\begin{table}[t]
\caption{Training on \cifar using a \cresnetl classifier on a single Nvidia V100 GPU. State-of-the-art robustness training methods significantly slow down training compared to standard training.}
\label{tab:std-training-comp}
\begin{center}
\begin{small}
\begin{sc}
\begin{tabular}{@{}lc@{}}
\toprule
\textbf{Method} & \textbf{Training Slowdown Factor} \\
\midrule
\smoothadv & $46.20 \times$ \\
\macer & $20.86 \times$ \\
\smoothmix & $4.97 \times$ \\
\bottomrule
\end{tabular}
\end{sc}
\end{small}
\end{center}
\vspace{-2em}
\end{table}

\subsection{Transferring Certified Robustness} \label{sec:xfer_cert_rob}
From Equation~\ref{eq:certified_radius}, it follows that training the base classifier to maximize $P_{\eta}(f_{\theta}(x + \eta) = y)$ for any given input $x$ will result in the maximization of the certified radius associated with the smooth classifier, provided Equation~\ref{eq:rs_cond} is satisfied. Thus, for the base classifier $f_{\theta}(x)$, our goal is to maximize the following quantity over the training set:

\vspace{-1em}
\begin{align} \label{eq:prob_sum}
    \sum_{i=1}^{n} \mathds{E}_{\eta} \mathbf{1}[f_{\theta}(x_i + \eta) = y_i] \approx \sum_{i=1}^{n} \mathds{E}_{\eta} [z_{\theta}^{y_i}(x_i + \eta)]
\end{align}
In the above equation, like prior works~\cite{cohen2019certified,salman2019provably,zhai2020macer}, we leverage the fact that the softmax output of a classifier can be treated as a continuous and differentiable approximation of its $\arg\max$ output. Methods like \smoothadv~\cite{salman2019provably}, \macer~\cite{zhai2020macer} and \smoothmix~\cite{jeong2021smoothmix} that target $\ell_2$ certifiable robustness propose training objectives that maximize this term.

Now, suppose we have a pre-trained base classifier $f_{\phi}$. It follows that $\mathds{E}_{\eta} [z_{\phi}^y(x + \eta)] \geq 0$. Through straightforward algebraic manipulations (see Appendix~\ref{app:crt_proof}), we derive the following lower bound:
\begin{align} \label{eq:crt_ineq}
\begin{split}
    \sum_{i=1}^{n} \mathds{E}_{\eta} [z_{\theta}^{y_i}(x_i + \eta)] &\geq -\sum_{i=1}^{n} \mathds{E}_{\eta} [z_{\phi}^{y_i}(x_i + \eta) - z_{\theta}^{y_i}(x_i + \eta)]
\end{split}
\end{align}
That is to say that, for a given input $x_i$, if we minimize the difference between the softmax outputs of the teacher and the student ($f_{\phi}$ and $f_{\theta}$) corresponding to the correct label $y_i$, we will maximize Equation~\ref{eq:prob_sum} for the student. However, to ensure that the student has a non-trivial certified radius, we must also ensure that Equation~\ref{eq:rs_cond} is satisfied. If we assume that Equation~\ref{eq:rs_cond} holds true for the teacher (\ie the base classifier of a certifiably robust smooth classifier), this condition can also be achieved for the student by matching the overall softmax output of the student to that of the teacher.

\subsection{\syslong (\sys)}
Based on the previous discussion, we now describe our method for training a certifiably robust classifier through knowledge transfer. First, we obtain a pre-trained base classifier $f_\phi$, which has been trained using a randomized smoothing based robustness training method as this maximizes ${\mathds{E}_{\eta} [z_{\phi}^y(x + \eta)]}$. Next, we use $f_\phi$ as a teacher to train a new student base classifier $f_\theta$. The student is trained to match the output of the teacher. In doing so, we can maximize the certified radius of the associated smooth classifier $g_\theta$ (Equation~\ref{eq:crt_ineq}), as well as ensure that Equation~\ref{eq:rs_cond} is satisfied. We describe our implementation in Algorithm~\ref{alg:pseudocode}. Given a batch of inputs, we first perturb them with additive Gaussian noise. Next, we compute the $\ell_2$ distance between the student and the teacher's outputs for these Gaussian perturbed inputs. This distance serves as our loss function, and we update the parameters of the student to minimize this loss.
At test time, the classifier $f_\theta$ is converted to its smooth version $g_\theta$ following Definition~\ref{def:smooth}.

\begin{algorithm}[t!]
    \caption{\syslong (\sys)} \label{alg:pseudocode}
    \begin{algorithmic}[1]
    \STATE \textbf{Input:}\xspace Training data distribution $\mathcal{D}$, certifiably robust teacher base classifier $f_\phi$, noise level $\sigma$, total training iterations $\mathcal{T}$, learning rate $\alpha$
    \STATE \textbf{Output:}\xspace Certifiably robust student base classifier $f_\theta$
    \STATE $\theta \gets$ random initialization
    \STATE $i \gets 0$
    \WHILE{$i < \mathcal{T}$}
        \STATE From $\mathcal{D}$, sample a batch of inputs $\{x_1,x_2,\cdots,x_n\}$.
        \STATE From $\mathcal{N}(0,\sigma^2 I)$, generate a batch of Gaussian noise samples $\{\eta_1,\eta_2,\cdots,\eta_n\}$.
        \STATE $l_i \gets \frac{1}{n} \sum_{j=1}^n \| z_\phi(x_j + \eta_j) - z_\theta(x_j + \eta_j) \|_2$
        \STATE $\theta \gets \theta - \alpha \cdot \nabla_{\theta} l_i$
        \STATE $i\gets i+1$
    \ENDWHILE
    \end{algorithmic}
\end{algorithm}

\subsection{Prior Works on Robustness Transfer}
Several prior works have examined transferring adversarial robustness between classifiers, but these works have been limited to transferring empirical rather than certified robustness~\cite{chan2020thinks,Goldblum_2020,Ilyas2019AdversarialEA,zhu2021reliable,zi2021revisiting}. Of note is the work by  Goldblum~\etal~\cite{Goldblum_2020} in which they combine adversarial training~\cite{madry2018towards} with knowledge distillation~\cite{hinton2015distilling}. They show that distilling knowledge from a large network to a small network improves its empirical robustness as compared to training the small network on its own, but their method makes no effort to improve the computational cost of adversarial training.

\section{Evaluation} \label{sec:eval}
Our goal is to improve the usability of randomized smoothing based robustness training methods. In this section, we demonstrate how \sys enables the reuse of an existing certifiably robust classifier to train new certifiably robust classifiers at significantly reduced training cost compared to prior methods. In our first experiment, we train a \cresnetl classifier with a state-of-the-art method, \ie \smoothmix~\cite{jeong2021smoothmix}), and use \sys to transfer its robustness to train several newer generation classifiers. In a second experiment, we recursively use \sys to train a newer generation classifier using the previous generation classifier that was also trained using \sys. In each experiment, we compare the certified robustness of classifier trained using \sys against a classifier trained using \smoothmix (Section~\ref{sec:cert_rob}). We find that classifiers trained using \sys are similarly robust as when trained using \smoothmix but only require a fraction of training time (Section~\ref{sec:training_time}). Our main results are generated using the \cifar dataset~\cite{krizhevsky2009learning}, but we also demonstrate the effectiveness of \sys on \imagenet~\cite{deng2009imagenet} (Section~\ref{sec:imagenet}). Both these datasets are open-source and free for non-commercial use.

\paratitle{Architectures} We use several popular DNN architectures that were proposed to either improve upon the visual recognition performance of the previous generation architectures or preserve performance while requiring significantly fewer parameters (or both). For the \cifar experiments, we use \cresnetl~\cite{he2016deep}, \resnext~\cite{xie2017aggregated}, \dla~\cite{yu2018deep}, and \regnet~\cite{radosavovic2020designing}.\footnote{We use the \cifar version of these architectures, code [MIT License]: \url{https://github.com/kuangliu/pytorch-cifar}.}


\paratitle{Training details}
All \smoothmix classifiers were trained using the code made available by the authors\footnote{\smoothmix code [MIT License]: \url{https://github.com/jh-jeong/smoothmix}} and the hyperparameters reported by them~~\cite{jeong2021smoothmix}. All \sys classifiers were trained using Stochastic Gradient Descent till convergence (200 epochs), with a batch size of 128. Further hyperparameter details are available in Appendix~\ref{app:hyperparams}. Unless specified, we report results for noise level $\sigma=0.25$ in the main paper. Additional results for higher noise levels $\sigma=0.5$ and $1.0$ are reported in Appendix~\ref{app:higher_noise}.

\paratitle{Evaluation Metrics} 
We report our results using two metrics. First, as done in prior work, we measure the \textbf{certified robustness} of a classifier based on (1)~the~\textit{certified test accuracy} at $\ell_2$ radius $r$~\footnote{Note that the certified accuracy at $r=0$ represents the clean accuracy of the smooth classifier.}, which is defined as the fraction of test set inputs that the smooth classifier classifies correctly within an $\ell_2$ ball of radius $r$ centered at each input, and (2)~\textit{average certified radius} (ACR), which is the average of the certified radius across all inputs in the test set: 
\begin{equation*}
\centering
ACR(g_\theta) = \frac{1}{n_{test}}\sum_{i=1}^{n_{test}} CR(g_\theta;x_i,y_i)    
\end{equation*}
On \cifar, we compute these metrics using the entire test set.
Second, we measure \textbf{training time} of a classifier based on the \textit{per-epoch time} and \textit{total training time}. The total training time is computed once the model's loss has converged. All classifiers were trained on the same machine with a single Nvidia Titan V GPU.

\subsection{Certified Robustness Comparison} \label{sec:cert_rob}

\paratitle{Standard \sys Training}
Given a \cresnetl classifier trained using \smoothmix, we transfer its robustness to several newer generation classifiers. We compare the certified robustness of these classifiers with their \smoothmix trained versions. The results are summarized in Table~\ref{tab:cifar_smix_0.25}. We observe that using \sys does not reduce the certified robustness of the trained classifier compared to training with \smoothmix. In fact, interestingly, \sys trained classifiers exhibit higher certified robustness compared to their \smoothmix baseline. Not only do \sys trained classifiers have higher ACR (improvement of $8.1\%$ in the best case), they also exhibit higher certified accuracy at different $\ell_2$ radii. Furthermore, \sys remains effective even as the generation gap between the student and the teacher increases. This implies that the same teacher can potentially be reused  indefinitely, amortizing the teacher's training cost to a constant. These results empirically validate our theoretical justification of \sys. Finally, we note that in Table~\ref{tab:cifar_smix_0.25}, the accuracy on clean inputs and ACR of \sys trained classifiers follow the same trend as in Figure~\ref{fig:model-evolution}, thus motivating the need for periodic model re-deployment to incorporate architectural improvements.

\begin{table*}[h]
\caption{The certified robustness of classifiers with different architectures trained on \cifar using \smoothmix~\cite{jeong2021smoothmix} and \sys. We use \sys to transfer the robustness of a \cresnetl trained using \smoothmix. We report certified test accuracy at different values of $\ell_2$ radius and the Average Certified Radius (ACR). The architectures are sorted chronologically based on published date. The noise level $\sigma$ is set to $0.25$.}
\vspace{1em}
\label{tab:cifar_smix_0.25}
\begin{center}
\begin{small}
\begin{sc}
\begin{tabular}{@{}lcccc|c@{}}
\toprule
\textbf{Architecture} & \textbf{0.00} & \textbf{0.25} & \textbf{0.50} & \textbf{0.75} & \textbf{ACR}\\
\midrule
\multicolumn{6}{c}{\smoothmix~\cite{jeong2021smoothmix}} \\
\midrule
\cresnetl~\cite{he2016deep} & 76.89 & 68.25 & 57.42 & 46.26 & 0.550 \\
\midrule
\resnext~\cite{xie2017aggregated} & 75.98 & 65.40 & 53.78 & 41.03 & 0.516 \\
\dla~\cite{yu2018deep} & 77.72 & 68.53 & 57.69 & 45.56 & 0.551 \\
\regnet~\cite{radosavovic2020designing} & 76.48 & 66.79 & 56.36 & 44.47 & 0.538 \\
\midrule
\multicolumn{6}{c}{\sys (\cresnetl Teacher)} \\
\midrule
\resnext~\cite{xie2017aggregated} & 77.57 & 69.00 & 58.31 & 47.16 & 0.558 \\
\dla~\cite{yu2018deep} & 77.31 & 68.91 & 58.26 & 46.34 & 0.554 \\
\regnet~\cite{radosavovic2020designing} & 77.89 & 69.57 & 59.36 & 47.28 & 0.564 \\
\bottomrule
\end{tabular}
\end{sc}
\end{small}
\end{center}
\vspace{-1em}
\end{table*}

\begin{table*}[h]
\caption{The certified robustness of classifiers with different architectures trained on \cifar using \sys recursively. We report certified test accuracy at different values of $\ell_2$ radius and the Average Certified Radius (ACR). Here, the previous generation classifier is used to train the current generation one. Chain length represents the number times \sys was used in training. The noise level $\sigma$ is set to $0.25$. \sys remains effective despite recursive use.}
\vspace{1em}
\label{tab:cifar_chain_0.25}
\begin{center}
\begin{small}
\begin{sc}
\begin{tabular}{@{}lccccc|c@{}}
\toprule
\textbf{Architecture} & \textbf{Chain Length} & \textbf{0.00} & \textbf{0.25} & \textbf{0.50} & \textbf{0.75} & \textbf{ACR}\\
\midrule
\resnext~\cite{xie2017aggregated} & 1 & 77.57 & 69.00 & 58.31 & 47.16 & 0.558 \\
\dla~\cite{yu2018deep} & 2 & 78.46 & 70.05 & 60.01 & 48.30 & 0.570 \\
\regnet~\cite{radosavovic2020designing} & 3 & 78.16 & 69.00 & 58.69 & 47.00 & 0.559 \\
\bottomrule
\end{tabular}
\end{sc}
\end{small}
\end{center}
\end{table*}

\paratitle{Recursive \sys Training} We now explore the effectiveness of \sys if it is used recursively, \ie the newest generation is trained using a \sys trained classifier from the previous generation as the teacher. We begin with a \cresnetl trained using \smoothmix. Then, all subsequent classifiers are trained using \sys recursively and report the results in Table~\ref{tab:cifar_chain_0.25}. The \textit{chain length} measures the number of times \sys was used. For example, the \dla network, with a chain length of 2, is the result of using \sys twice: once to transfer the \smoothmix trained \cresnetl network's performance to the \resnext network and once to transfer the \sys trained \resnext network's performance to the \dla network. We observe that the certified robustness of the resulting classifiers remains high even with recursive use of \sys. The empirical results are to be expected given our theoretical understanding of \sys: In order to train a robust student, we only require that the teacher is already robust (\ie satisfies the condition of Theorem~\ref{thm:rs}) \textit{irrespective of the training method used to achieve robustness}. Thus, we expect \sys to remain effective even at longer chain lengths. In Section~\ref{sec:teacher_training}, we will highlight the relationship between the teacher's training method and the robustness of a \sys trained student.

\subsection{Training Time Comparison}  \label{sec:training_time}
Having established that \sys effectively transfers certified robustness between classifiers, we now evaluate its training overhead. For comparison, we also evaluate the training overhead of \smoothmix. In Table~\ref{tab:cifar_time}, we report the per-epoch time and total time of training different architectures with each method. For brevity, we only compare the training time for standard \sys.\footnote{Recursive \sys differs in time by an insignificant factor due to forward pass through a different teacher.} We observe that the per-epoch time of \sys is significantly lower than \smoothmix. Similarly, when trained until convergence, the total training time of \sys is significantly lower. Across the three architectures that we run our experiments on, \sys achieves an average epoch time speedup of $10.75 \times$. Comparing overall training times, \sys speeds up training by, on average, $8.06 \times$. If we consider the real-world scenario where the model has to be periodically redeployed to incorporate architectural improvements, the cumulative training time using \smoothmix is 96.21 hours as each new architecture is trained from scratch. With \sys, the cumulative time is reduced to 11.70 hours representing a $87.84\%$ savings in costs associated with computational resources.

\paratitle{Teacher's availability} So far, we assumed the availability of a certifiably robust teacher (\cresnetl). We argue that this is a reasonable assumption as the amortized cost associated with the one-time training of a robust teacher is negligible across many generations of the model. Regardless, in Section~\ref{sec:teacher_unavailable}, we examine a scenario where the teacher is  unavailable. Under this scenario, we demonstrate how \sys can be used to speedup the training of \cresnetl for use as teacher.

\begin{table*}[h]
\caption{Training time statistics for \smoothmix and \sys. We report the mean and 95\% confidence interval computed over all training epochs. \sys is on average $8 \times$ faster than \smoothmix across all three architectures.}
\label{tab:cifar_time}
\begin{center}
\begin{small}
\begin{sc}
\resizebox{\columnwidth}{!}{%
\begin{tabular}{@{}lcccc@{}}
\toprule
\multirow{2}{*}[-0.4em]{\textbf{Architecture}} & \multicolumn{2}{c}{\textbf{\smoothmix~\cite{jeong2021smoothmix}}} & \multicolumn{2}{c}{\textbf{\sys (\cresnetl Teacher)}} \\ \cmidrule(lr){2-3} \cmidrule(lr){4-5}
 & \textbf{Epoch Time (s)} & \textbf{Total Time (h)} & \textbf{Epoch Time (s)} & \textbf{Total Time (h)} \\
\midrule
\cresnetl~\cite{he2016deep} & 455.55 $\pm$ 1.17 & 18.98 & - & - \\
\midrule
\resnext~\cite{xie2017aggregated} & 1085.09 $\pm$ 0.50 & 45.21 & 86.41 $\pm$ 0.11 & 4.80 \\
\dla~\cite{yu2018deep} & 854.41 $\pm$ 0.09 & 35.60 & 62.24 $\pm$ 0.40 & 3.46 \\
\regnet~\cite{radosavovic2020designing} & 369.42 $\pm$ 0.51 & 15.39 & 61.92 $\pm$ 0.30 & 3.44 \\
\bottomrule
\end{tabular}
}
\end{sc}
\end{small}
\end{center}
\vspace{-1em}
\end{table*}


\section{Discussion}
In this section, we address the standout concerns about \sys. The section layout is as follows: in Section~\ref{sec:teacher_unavailable}, we discuss the scenario in which a certifiably robust teacher is not readily available and demonstrate how \sys can still speed up robustness training; in Section~\ref{sec:teacher_training}, we examine how the method used to train the teacher affects the robustness of the student; in Section~\ref{sec:imagenet}, we study the scalability of \sys using the \imagenet dataset; in Section~\ref{sec:consistency_compare}, we compare \sys with a closely related prior work on fast certified robustness training, \ie Consistency regularization~\cite{jeong2020consistency}; in Section~\ref{sec:limitations}, we discuss the limitations of \sys; in Section~\ref{sec:impact}, we address the broader impact of \sys.

\subsection{Teacher Not Available} \label{sec:teacher_unavailable}
We've designed \sys under the assumption that a certifiably robust teacher is already available. However, even if a certifiably robust teacher is not available, \sys can still speed up training. Given a certifiable robust training method and a large network architecture, we can reduce the training overhead by robustly training a comparatively smaller network first. Then, we can use \sys to transfer the robustness of the small network to a larger network. In Table ~\ref{tab:cifar_small_to_large_0.25}, we present results for such a process. First, we trained a \cresnets network using \smoothmix, then we used \sys to train a \cresnetl network. We compare the robustness of a \cresnetl trained using this process with one trained using \smoothmix. As we can see, the \sys \cresnetl network has comparable robustness with the \smoothmix \cresnetl network. However, even when adding the teacher and student training times, \sys still speeds up training by approximately $2\times$ relative to \smoothmix.

\begin{table*}[h]
\caption{Certified robustness and total time of a \cresnetl classifier trained on \cifar using \smoothmix and \sys. For \sys, we train a \cresnets teacher first using \smoothmix and report total time as the time taken to train the teacher and the student. The noise level $\sigma$ is set to $0.25$. The \cresnetl trained using \sys achieves an ACR comparable to the \smoothmix \cresnetl while achieving a $\sim 2 \times$ speedup in total training time.}
\vspace{1em}
\label{tab:cifar_small_to_large_0.25}
\begin{center}
\begin{small}
\begin{sc}
\begin{tabular}{@{}lcccc|c|c}
\toprule
\textbf{Method} & \textbf{0.00} & \textbf{0.25} & \textbf{0.50} & \textbf{0.75} & \textbf{ACR} & \textbf{Total Time (h)}\\
\midrule
\smoothmix~\cite{jeong2021smoothmix} & 76.89 & 68.25 & 57.42 & 46.26 & 0.550 & 18.98 \\
\sys (\cresnets Teacher) & 75.68 & 67.20 & 56.30 & 44.83 & 0.540 & 10.07 \\
\bottomrule
\end{tabular}
\end{sc}
\end{small}
\end{center}
\vspace{-1em}
\end{table*}

\subsection{Teacher Training Method} \label{sec:teacher_training}
We train a \cresnets classifier using \macer~\cite{zhai2020macer}, \smoothadv~\cite{salman2019provably}, and \smoothmix~\cite{jeong2021smoothmix}. For \macer and \smoothadv training, we use the code made available by the authors\footnote{\macer code [No license available]: \url{https://github.com/RuntianZ/macer}}\footnote{\smoothadv code [MIT License]: \url{https://github.com/Hadisalman/smoothing-adversarial}} and the hyperparameters reported by them. Using \sys, we transfer the robustness of each teacher to a \cresnetl classifier. The results are reported in Table~\ref{tab:cifar_diff_teacher_0.25}.
For reference, we also report robustness of a \cresnetl network trained independently using the chosen robustness training methods.
Overall, we observe a slight variation in the robustness of the \sys trained networks depending on the teachers training method. Based on Equation \ref{eq:crt_ineq}, this is expected as maximizing the teacher's performance will in turn maximize the performance of the student. Our empirical results align with this reasoning: the \macer teacher was the least robust of the three methods, and its student is similarly the least robust of the students. However, in all cases, the \sys trained network obtained certified robustness comparable to its teacher.

\begin{table}[h]
\caption{For \cifar dataset, certified robustness achieved on training the \sys teacher (\cresnets) with different methods. The student classifier is \cresnetl. For reference, we also report robustness of \cresnetl trained independently using chosen methods. The noise level $\sigma$ is set to $0.25$. Students attain comparable ACR to their respective teachers.}
\label{tab:cifar_diff_teacher_0.25}
\begin{center}
\begin{small}
\begin{sc}
\begin{tabular}{@{}lcclc@{}}
\toprule
\multicolumn{2}{c}{\textbf{Teacher (\cresnets)}} & & \multicolumn{2}{c}{\textbf{Student (\cresnetl)}} \\ \cmidrule(lr){1-2} \cmidrule(lr){4-5}
\textbf{Training Method} & \textbf{ACR} & & \textbf{Training Method} & \textbf{ACR} \\
\midrule
\smoothadv~\cite{salman2019provably} & 0.531 & & \sys & 0.519 \\
\macer~\cite{zhai2020macer} & 0.507 & $\rightarrow$ & \sys & 0.528 \\
\smoothmix~\cite{jeong2021smoothmix} & 0.522 & & \sys & 0.540 \\
\midrule
 & & & \smoothadv~\cite{salman2019provably} & 0.547\\
 \multicolumn{2}{c}{Student trained directly} & & \macer~\cite{zhai2020macer} & 0.531 \\
 & & & \smoothmix~\cite{jeong2021smoothmix} & 0.550 \\
\bottomrule
\end{tabular}
\end{sc}
\end{small}
\end{center}
\vspace{-1em}
\end{table}


\subsection{Scalability} \label{sec:imagenet}
Here, we study the effectiveness of \sys on a large-scale dataset, \ie \imagenet. For this purpose, we train \iresnets classifiers using three certified robustness training methods (\macer, \smoothadv, and \smoothmix). Next, we transfer their robustness to a \iresnetl classifier. The results were generated on a 500 sample test set (following prior works~\cite{salman2019provably,zhai2020macer,jeong2021smoothmix}) and are summarized in Table~\ref{tab:imagenet_rob}.
For reference, we also report robustness of a \iresnetl network trained independently using the chosen robustness training methods.
In all cases, we observe that students achieve certified robustness comparable to their respective teachers.
Therefore, \sys remains effective even on a more complex dataset.

\begin{table}[h]
\caption{\imagenet results using \sys and three robustness training methods. We report both the ACR of the \iresnets teacher and its \iresnetl student. For reference, we also report robustness of \iresnetl trained independently using chosen methods. The noise level $\sigma$ is set to $0.5$. Students attain comparable ACR to their respective teachers.}
\label{tab:imagenet_rob}
\begin{center}
\begin{small}
\begin{sc}
\begin{tabular}{@{}lcclc@{}}
\toprule
\multicolumn{2}{c}{\textbf{Teacher (\iresnets)}} & & \multicolumn{2}{c}{\textbf{Student (\iresnetl)}} \\ \cmidrule(lr){1-2} \cmidrule(lr){4-5}
\textbf{Training Method} & \textbf{ACR} & & \textbf{Training Method} & \textbf{ACR} \\
\midrule
\smoothadv~\cite{salman2019provably} & 0.684 & & \sys & 0.684 \\
\macer~\cite{zhai2020macer} & 0.574 & $\rightarrow$ & \sys & 0.576 \\
\smoothmix~\cite{jeong2021smoothmix} & 0.653 & & \sys & 0.661 \\
\midrule
 & & & \smoothadv~\cite{salman2019provably} & 0.820 \\
 \multicolumn{2}{c}{Student trained directly} & & \macer~\cite{zhai2020macer} & 0.653 \\
 & & & \smoothmix~\cite{jeong2021smoothmix} & 0.799 \\
\bottomrule
\end{tabular}
\end{sc}
\end{small}
\end{center}
\end{table}

\newpage

\subsection{Comparison with Consistency Regularization~\cite{jeong2020consistency}} \label{sec:consistency_compare}
In Section \ref{sec:eval}, we compared \sys against \smoothmix as it has state-of-the-art ACR. However, another closely related work was recently published by Jeong~\&~Shin~\cite{jeong2020consistency}, which shows state-of-the-art ACR and potential training time improvements. They proposed a consistency regularization loss that improves the certified robustness of smooth classifiers by enforcing the base classifier's soft outputs to be consistent across multiple noisy copies of a given input. Therefore, their additional computational overhead scales linearly with the number of noisy samples used to compute the consistency loss. With respect to computational overhead, \sys adds only one forward pass, \ie the pass through the teacher. When paired with Gaussian data augmentation training, their regularization loss significantly improves the certified robustness of a smooth classifier. By applying their regularization loss over only two noisy copies of the input, they can achieve better certified robustness than prior state-of-the-art robustness training methods like \macer~\cite{zhai2020macer} and \smoothadv~\cite{salman2019provably}.  

The key difference between \sys and consistency regularization is in the training overhead when combined with other state-of-the-art certified training methods. Consistency regularization augments classifier training with an additional loss term. Therefore, their training overhead is dominated by the training method selected. In their experiments, they focused on Gaussian data augmentation, which adds little to no training overhead relative to standard training. However, if a more computationally intensive method was selected (\eg \macer), they remark their training overhead would dramatically increase. With respect to \sys, if a teacher is available (\ie a previous generation model), the overhead of \sys is agnostic to the training method. If it is not available, we demonstrated in Section \ref{sec:teacher_unavailable}, that \sys can still greatly reduce training overhead. For interested readers, we include results for transferring robustness from a teacher trained using Consistency regularization in Appendix~\ref{app:consistency}.

\newpage

\subsection{Limitations} \label{sec:limitations}
In this paper, we use probabilistic certified robustness methods as they rely on Theorem~\ref{thm:rs} and, thus, are designed to maximize the certified radius (Equation~\ref{eq:certified_radius}). We found that deterministic methods (\eg CROWN-IBP~\cite{zhang2019towards}) impose a stricter training requirement on the base teacher classifier. For a given input, deterministic training methods require the base classifier to be correct for all inputs within the $\ell_2$-norm ball, rather than simply be likely to correctly classify inputs within the $\ell_2$-norm ball. This restriction lowers the potential ACR of the smooth teacher classifier, which also lowers the ACR of the student trained using \sys. For example, when using CROWN-IBP~\cite{zhang2019towards} to train a ResNeXt base classifier, the ACR for the corresponding smooth classifier is only 0.064. When transferring the robustness of this ResNeXt classifier to a WideResNet34-10 student, we get an ACR of 0.065.

Additionally, we note that the classifier architectures we present in the paper are restricted to CNNs. Recently, a new class of transformer-based image classifiers~\cite{dosovitskiy2020image,liu2021swin,dai2021coatnet} have been proposed that show improved performance over CNN classifiers. We briefly studied the effectiveness of \sys when transferring robustness between CNN and transformer architectures using \vit~\cite{dosovitskiy2020image} and present the results in Appendix~\ref{app:vit_results}, but further exploration is needed.
Finally, \sys has only been studied using the $\ell_2$ norm and image data due to the limitations of current certified robustness training methods.

\subsection{Broader Impacts} \label{sec:impact}
As we have shown, our work improves the efficiency of training certifiably robust classifiers, in an effort to improve the security of AI-powered systems. Beyond the broad negative societal impacts of machine learning, we are not aware of any impacts specific to our work.

\section{Conclusion}

In this paper, we proposed the first general-purpose framework to speed up the training of certifiably robust classifiers using knowledge transfer and randomized smoothing. Our proposed method, \syslong (\sys) enables transferring the certified robustness of a classifier to another classifier at a cost comparable to standard training. We provided a theoretical understanding of \sys and provided empirical evidence of its effectiveness. On \cifar, we showed that across several generations of classifier architectures, \sys trained classifiers $8 \times$ faster than when using a state-of-the-art training method, while achieving comparable or better certified robustness. Furthermore, \sys can reduce the training overhead of certified robustness training methods even when an initial robust classifier is not present. The use of machine learning in security and safety critical environments motivates a need for models with certifiably robust performance, but the training overhead of existing certified robustness training methods inhibits their usability. Our work addresses this issue, especially for commercial applications where periodical model re-deployment is inevitable.



\section*{Acknowledgement}
This work was supported by the Office of Naval Research under grants N00014-20-1-2858 and N00014-22-1-2001,  Air Force Research Lab under grant FA9550-22-1-0029, and NVIDIA 2018 GPU Grant. Any opinions, findings, or conclusions expressed in this material are those of the authors and do not necessarily reflect the views of the sponsors.

\bibliographystyle{plain}
\bibliography{refs}

\newpage

\section*{Checklist}


\begin{enumerate}

\item For all authors...
\begin{enumerate}
  \item Do the main claims made in the abstract and introduction accurately reflect the paper's contributions and scope?
    \answerYes{See Section~\ref{sec:eval}.}
  \item Did you describe the limitations of your work?
    \answerYes{See Section~\ref{sec:limitations}.}
  \item Did you discuss any potential negative societal impacts of your work?
    \answerNo{In this paper, we propose a method to improve the usability of certified robustness training methods. Furthermore, beyond the broad negative societal impacts that results for ML, we are unaware of any specific to our work.}
  \item Have you read the ethics review guidelines and ensured that your paper conforms to them?
    \answerYes{We have read the ethics review guidelines and acknowledge that our paper conforms to them.}
\end{enumerate}

\item If you are including theoretical results...
\begin{enumerate}
  \item Did you state the full set of assumptions of all theoretical results?
    \answerYes{See Section~\ref{sec:background} for foundational theoretical results from prior work and Section~\ref{sec:method} for the \sys specific results.}
        \item Did you include complete proofs of all theoretical results?
    \answerYes{The complete theoretical justification for \sys is included in Appendix~\ref{app:crt_proof}. For proofs for other foundational results such as Theorem \ref{thm:rs}, see the respective paper.}
\end{enumerate}

\item If you ran experiments...
\begin{enumerate}
  \item Did you include the code, data, and instructions needed to reproduce the main experimental results (either in the supplemental material or as a URL)?
    \answerYes{See Appendix~\ref{app:training_details}.}
  \item Did you specify all the training details (e.g., data splits, hyperparameters, how they were chosen)?
    \answerYes{See Appendix~\ref{app:hyperparams}.}
        \item Did you report error bars (e.g., with respect to the random seed after running experiments multiple times)?
    \answerYes{We include 95\% confidence interval for our timing results. However, computing error bars for certified accuracy and ACR requires training multiple classifiers and performing the costly process of certification multiple times. Additionally, we observed that most prior works do not compute such error bars.}
        \item Did you include the total amount of compute and the type of resources used (e.g., type of GPUs, internal cluster, or cloud provider)?
    \answerNo{While in Sections~\ref{sec:training_time}~and~\ref{sec:teacher_unavailable}, we include the total training time results to support our claims as well as the GPU model. We do not include such information for all experiments.}
\end{enumerate}

\item If you are using existing assets (e.g., code, data, models) or curating/releasing new assets...
\begin{enumerate}
  \item If your work uses existing assets, did you cite the creators?
    \answerYes{}
  \item Did you mention the license of the assets?
    \answerYes{When specified by the asset creator, we included the license.}
  \item Did you include any new assets either in the supplemental material or as a URL?
    \answerYes{See Appendix~\ref{app:code}.}
  \item Did you discuss whether and how consent was obtained from people whose data you're using/curating?
    \answerNA{}
  \item Did you discuss whether the data you are using/curating contains personally identifiable information or offensive content?
    \answerNA{}
\end{enumerate}

\item If you used crowdsourcing or conducted research with human subjects...
\begin{enumerate}
  \item Did you include the full text of instructions given to participants and screenshots, if applicable?
    \answerNA{}
  \item Did you describe any potential participant risks, with links to Institutional Review Board (IRB) approvals, if applicable?
    \answerNA{}
  \item Did you include the estimated hourly wage paid to participants and the total amount spent on participant compensation?
    \answerNA{}
\end{enumerate}

\end{enumerate}

\newpage

\appendix
\section{Justification for \sys Training Objective} \label{app:crt_proof}

Following from the discussion in Section~3.1, we want to maximize $\mathds{E}_{\eta} [z_{\theta}^y(x + \eta)]$. Given a non-negative random variable $t=\mathds{E}_{\eta} [z_{\phi}^y(x + \eta)]$, we have
\begin{align*}
    \mathds{E}_{\eta} [z_{\theta}^y(x + \eta)] \geq \mathds{E}_{\eta} [z_{\theta}^y(x + \eta)] - t
\end{align*}
Here $z_\phi$ is the soft output associated with classifier $f_\phi$ that has been trained independently of $f_\theta$. Therefore, we have
\begin{align*} \label{eq:crt_proof}
\begin{split}
    \mathds{E}_{\eta} [z_{\theta}^y(x + \eta)] &\geq \mathds{E}_{\eta} [z_{\theta}^y(x + \eta)] - \mathds{E}_{\eta} [z_{\phi}^y(x + \eta)] \\
    &\geq \mathds{E}_{\eta} [z_{\theta}^y(x + \eta) - z_{\phi}^y(x + \eta)] \\
    &\geq -\mathds{E}_{\eta} [z_{\phi}^y(x + \eta) - z_{\theta}^y(x + \eta)]
\end{split}
\end{align*}
This implies that maximizing $\mathds{E}_{\eta} [z_{\theta}^y(x + \eta)]$ is equivalent to minimizing $\mathds{E}_{\eta} [z_{\phi}^y(x + \eta) - z_{\theta}^y(x + \eta)]$.

\section{Additional Results} \label{app:additional_results}

\subsection{Higher Noise Level} \label{app:higher_noise}
In the main paper, we conduct experiments on \cifar using noise level $\sigma=0.25$ only. Here, we report our main set of results on \cifar (Table~\ref{tab:cifar_chain_0.25}) using higher $\sigma$ values. In Table~\ref{tab:cifar_smix_0.5}, we report results using $\sigma=0.5$ and in Table~\ref{tab:cifar_smix_1.0}, we report results using $\sigma=1.0$.

\subsection{Using ViT~\cite{dosovitskiy2020image}} \label{app:vit_results}
In the main paper, we used Convolutional Neural Network (CNN) based architectures. However, there is another recently developed class of architectures that use Transformers for the task of image classification ~\cite{dosovitskiy2020image,liu2021swin,dai2021coatnet}. In Table~\ref{tab:cifar_vit}, we present results measuring the effectiveness of \sys in transferring robustness from \cresnetl (a CNN-based classifier) to \vit~\cite{dosovitskiy2020image} (a Transformer-based classifier). For comparison, we also report results obtained on training \vit with \smoothmix. \sys trained \vit classifiers perform comparable or better than their \smoothmix counterparts.

\subsection{Training \smoothmix Classifiers with \sys Hyperparameters} \label{app:higher_epochs}
For generating results using prior methods, we strictly adhere to the hyperparameters reported by them. However, the hyperparameters that we use for \sys training is different than the ones used by prior methods (see Table~\ref{tab:train_hyperparams}). In Table~\ref{tab:cifar_smix_200}, we report results obtained on using \sys training hyperparameters with \smoothmix. We note that there is not a significant difference in the robustness achieved using the two sets of hyperparameters.

\subsection{Training Teacher with Consistency Regularization~\cite{jeong2020consistency}} \label{app:consistency}
For our main set of results on \cifar, we focused on \smoothmix. In Section~\ref{sec:consistency_compare}, we discussed Consistency Regularization recently proposed by Jeong~\&~Shin~\cite{jeong2020consistency} as another method to attain certifiably robust classifiers with at a cost comparable to standard training depending on the setting. Here, we show results when Consistency Regularization is used to train the teacher classifier in Tables~\ref{tab:cifar_cons_0.25}~and~\ref{tab:cifar_chain_cons_0.25}. For comparison, in Table~\ref{tab:cifar_cons_0.25}, we also report the results for training the classifiers using Consistency Regularization. As with other training methods, \sys is effective in transferring the robustness of the teacher classifier irrespective of the training method.

\subsection{\imagenet Results} \label{app:extra_imagenet}
In Table~\ref{tab:imagenet_extra}, we present an additional \imagenet result using a different student-teacher pair. We note that, as in Table \ref{tab:imagenet_rob}, \sys remains effective.

\subsection{Gaussian Data Augmentation Baseline} \label{app:gda_results}
Along with the theoretical framework for creating certifiably robust image classifiers using randomized smoothing, Cohen~\etal~\cite{cohen2019certified} also proposed a simple yet effective method for training classifiers with high certified robustness within this framework. This method involves training the base classifier with Gaussian data augmentation. To date, this method remains the fastest way to train classifiers with non-trivial certified robustness using the randomized smoothing framework. However, this method is not as sophisticated as the more recently proposed methods, and so yields much poorer certified robustness than them. Since our work is focused at accelerating certified robustness training, in Table~\ref{tab:cifar_gda} we include the certified robustness of training time results for Gaussian data augmentation baseline for a more thorough comparison.


\begin{table*}[h]
\caption{The certified robustness of classifiers with different architectures trained on \cifar using \smoothmix~\cite{jeong2021smoothmix} and \sys. We use \sys to transfer the robustness of a \cresnetl trained using \smoothmix. We report certified test accuracy at different values of $\ell_2$ radius and the Average Certified Radius (ACR). The noise level $\sigma$ is set to $0.5$.}
\vspace{1em}
\label{tab:cifar_smix_0.5}
\begin{center}
\begin{small}
\begin{sc}
\resizebox{0.9\columnwidth}{!}{%
\begin{tabular}{@{}lcccccccc|c@{}}
\toprule
\textbf{Architecture} & \textbf{0.00} & \textbf{0.25} & \textbf{0.50} & \textbf{0.75} & \textbf{1.00} & \textbf{1.25} & \textbf{1.50} & \textbf{1.75} & \textbf{ACR}\\
\midrule
\multicolumn{10}{c}{\smoothmix~\cite{jeong2021smoothmix}} \\
\midrule
\cresnetl~\cite{he2016deep} & 65.01 & 57.71 & 49.99 & 42.74 & 35.98 & 29.43 & 23.52 & 17.33 & 0.725 \\
\midrule
\resnext~\cite{xie2017aggregated} & 63.90 & 56.81 & 48.80 & 40.79 & 33.29 & 27.24 & 20.80 & 14.60 & 0.687 \\
\dla~\cite{yu2018deep} & 65.76 & 58.47 & 51.16 & 43.97 & 37.16 & 30.50 & 23.97 & 18.02 & 0.742 \\
\regnet~\cite{radosavovic2020designing} & 64.75 & 57.48 & 49.96 & 42.57 & 35.23 & 28.79 & 22.78 & 16.56 & 0.716 \\
\midrule
\multicolumn{10}{c}{\sys (\cresnetl Teacher)} \\
\midrule
\resnext~\cite{xie2017aggregated} & 64.89 & 57.81 & 50.63 & 43.39 & 36.49 & 30.07 & 23.92 & 17.40 & 0.732 \\
\dla~\cite{yu2018deep} & 65.23 & 58.33 & 51.23 & 44.04 & 37.09 & 30.47 & 24.39 & 18.37 & 0.743 \\
\regnet~\cite{radosavovic2020designing} & 65.35 & 58.18 & 50.87 & 43.74 & 36.83 & 30.33 & 24.17 & 18.04 & 0.739 \\
\bottomrule
\end{tabular}
}
\end{sc}
\end{small}
\end{center}
\vspace{-1em}
\end{table*}

\begin{table*}[h]
\caption{The certified robustness of classifiers with different architectures trained on \cifar using \smoothmix~\cite{jeong2021smoothmix} and \sys. We use \sys to transfer the robustness of a \cresnetl trained using \smoothmix. We report certified test accuracy at different values of $\ell_2$ radius and the Average Certified Radius (ACR). The noise level $\sigma$ is set to $1.0$.}
\vspace{1em}
\label{tab:cifar_smix_1.0}
\begin{center}
\begin{small}
\begin{sc}
\resizebox{\textwidth}{!}{%
\begin{tabular}{@{}lcccccccccc|c@{}}
\toprule
\textbf{Architecture} & \textbf{0.00} & \textbf{0.25} & \textbf{0.50} & \textbf{0.75} & \textbf{1.00} & \textbf{1.25} & \textbf{1.50} & \textbf{1.75} & \textbf{2.00} & \textbf{2.25} & \textbf{ACR}\\
\midrule
\multicolumn{12}{c}{\smoothmix~\cite{jeong2021smoothmix}} \\
\midrule
\cresnetl~\cite{he2016deep} & 47.93 & 43.46 & 38.43 & 33.24 & 29.05 & 25.05 & 21.55 & 18.12 & 15.19 & 12.40 & 0.730 \\
\midrule
\resnext~\cite{xie2017aggregated} & 46.81 & 41.38 & 36.27 & 31.46 & 27.15 & 22.85 & 19.36 & 16.27 & 13.24 & 10.49 & 0.667 \\
\dla~\cite{yu2018deep} & 49.40 & 44.13 & 38.86 & 33.94 & 29.27 & 25.14 & 21.63 & 18.38 & 15.29 & 12.49 & 0.738 \\
\regnet~\cite{radosavovic2020designing} & 47.32 & 42.87 & 38.25 & 33.33 & 29.17 & 25.80 & 21.99 & 18.51 & 15.66 & 13.05 & 0.743 \\
\midrule
\multicolumn{12}{c}{\sys (\cresnetl Teacher)} \\
\midrule
\resnext~\cite{xie2017aggregated} & 48.03 & 43.41 & 38.56 & 33.15 & 28.92 & 25.29 & 21.43 & 18.40 & 15.03 & 12.20 & 0.728 \\
\dla~\cite{yu2018deep} & 48.38 & 43.67 & 38.82 & 33.76 & 29.4 & 25.53 & 21.94 & 18.71 & 15.32 & 12.61 & 0.741 \\
\regnet~\cite{radosavovic2020designing} & 48.18 & 43.53 & 38.68 & 33.62 & 29.30 & 25.44 & 21.86 & 18.45 & 15.29 & 12.46 & 0.735 \\
\bottomrule
\end{tabular}
}
\end{sc}
\end{small}
\end{center}
\vspace{-1em}
\end{table*}

\begin{table*}[h]
\caption{The certified robustness of a \vit classifier trained on \cifar using \smoothmix~\cite{jeong2021smoothmix} and \sys for different $\sigma$ values (\ie noise levels). We use \sys to transfer the robustness of a \cresnetl trained using \smoothmix. We report certified test accuracy at different values of $\ell_2$ radius and the Average Certified Radius (ACR).}
\label{tab:cifar_vit}
\begin{center}
\begin{small}
\begin{sc}
\resizebox{\textwidth}{!}{%
\begin{tabular}{@{}l|cccccccccc|c@{}}
\toprule
\textbf{$\sigma$} & \textbf{0.00} & \textbf{0.25} & \textbf{0.50} & \textbf{0.75} & \textbf{1.00} & \textbf{1.25} & \textbf{1.50} & \textbf{1.75} & \textbf{2.00} & \textbf{2.25} & \textbf{ACR}\\
\midrule
\multicolumn{12}{c}{\smoothmix~\cite{jeong2021smoothmix}} \\
\midrule
0.25 & 69.38 & 56.65 & 42.34 & 28.47 & 0.00 & 0.00 & 0.00 & 0.00 & 0.00 & 0.00 & 0.415 \\
0.50 & 50.56 & 43.87 & 37.20 & 30.55 & 24.55 & 19.41 & 14.78 & 10.43 & 0.00 & 0.00 & 0.515 \\
1.00 & 35.95 & 31.55 & 27.63 & 23.84 & 20.36 & 17.09 & 14.16 & 11.87 & 9.85 & 8.05 & 0.509 \\
\midrule
\multicolumn{12}{c}{\sys (\cresnetl Teacher)} \\
\midrule
0.25 & 69.63 & 56.60 & 42.29 & 28.45 & 0.00 & 0.00 & 0.00 & 0.00 & 0.00 & 0.00 & 0.415 \\
0.50 & 60.64 & 53.62 & 46.07 & 39.49 & 32.13 & 25.59 & 19.76 & 14.36 & 0.00 & 0.00 & 0.653 \\
1.00 & 41.76 & 37.19 & 32.55 & 28.43 & 24.67 & 20.89 & 17.48 & 14.91 & 12.29 & 9.79 & 0.610 \\
\bottomrule
\end{tabular}
}
\end{sc}
\end{small}
\end{center}
\end{table*}

\begin{table*}[h]
\caption{The certified robustness of classifiers with different architectures trained on \cifar using \smoothmix~\cite{jeong2021smoothmix}. Here, we use the same training hyperparameters as the ones we used to train \sys classifiers (see Table~\ref{tab:train_hyperparams}). We report certified test accuracy at different values of $\ell_2$ radius and the Average Certified Radius (ACR). The architectures are sorted chronologically based on published date. The noise level $\sigma$ is set to $0.25$.}
\vspace{1em}
\label{tab:cifar_smix_200}
\begin{center}
\begin{small}
\begin{sc}
\begin{tabular}{@{}lcccc|c@{}}
\toprule
\textbf{Architecture} & \textbf{0.00} & \textbf{0.25} & \textbf{0.50} & \textbf{0.75} & \textbf{ACR}\\
\midrule
\cresnetl~\cite{he2016deep} & 78.22 & 69.23 & 58.71 & 46.61 & 0.559 \\
\resnext~\cite{xie2017aggregated} & 77.22 & 66.72 & 55.06 & 42.43 & 0.528 \\
\dla~\cite{yu2018deep} & 78.07 & 69.37 & 58.61 & 46.70 & 0.559 \\
\regnet~\cite{radosavovic2020designing} & 77.39 & 68.06 & 57.25 & 45.44 & 0.547 \\
\bottomrule
\end{tabular}
\end{sc}
\end{small}
\end{center}
\end{table*}

\begin{table*}[h]
\caption{The certified robustness of classifiers with different architectures trained on \cifar using Consistency Regularization~\cite{jeong2020consistency} and \sys. We use \sys to transfer the robustness of a \cresnetl trained using Consistency Regularization. We report certified test accuracy at different values of $\ell_2$ radius and the Average Certified Radius (ACR). The noise level $\sigma$ is set to $0.25$.}
\vspace{1em}
\label{tab:cifar_cons_0.25}
\begin{center}
\begin{small}
\begin{sc}
\begin{tabular}{@{}lcccc|c@{}}
\toprule
\textbf{Architecture} & \textbf{0.00} & \textbf{0.25} & \textbf{0.50} & \textbf{0.75} & \textbf{ACR}\\
\midrule
\multicolumn{6}{c}{Consistency Regularization~\cite{jeong2020consistency}} \\
\midrule
\cresnetl~\cite{he2016deep} & 75.89 & 68.02 & 58.04 & 46.84 & 0.552 \\
\midrule
\resnext~\cite{xie2017aggregated} & 74.99 & 65.96 & 55.46 & 43.01 & 0.528 \\
\dla~\cite{yu2018deep} & 76.76 & 67.88 & 57.36 & 45.86 & 0.547 \\
\regnet~\cite{radosavovic2020designing} & 75.48 & 66.76 & 55.96 & 43.76 & 0.534 \\
\vit~\cite{dosovitskiy2020image} & 60.52 & 52.17 & 42.87 & 34.08 & 0.416 \\
\midrule
\multicolumn{6}{c}{\sys (\cresnetl Teacher)} \\
\midrule
\resnext~\cite{xie2017aggregated} & 75.72 & 68.46 & 58.85 & 47.30 & 0.557 \\
\dla~\cite{yu2018deep} & 76.61 & 69.18 & 59.59 & 48.35 & 0.565 \\
\regnet~\cite{radosavovic2020designing} & 76.18 & 68.42 & 58.85 & 47.64 & 0.558 \\
\vit~\cite{dosovitskiy2020image} & 73.00 & 64.50 & 53.69 & 42.33 & 0.515 \\
\bottomrule
\end{tabular}
\end{sc}
\end{small}
\end{center}
\end{table*}

\begin{table*}[h]
\caption{The certified robustness of classifiers with different architectures trained on \cifar using \sys recursively. The initial classifier was trained using Consistency Regularization~\cite{jeong2020consistency}. We report certified test accuracy at different values of $\ell_2$ radius and the Average Certified Radius (ACR). Here, the previous generation classifier is used to train the current generation one. Chain length represents the number times \sys was used in training. The noise level $\sigma$ is set to $0.25$. \sys remains effective despite recursive use.}
\vspace{1em}
\label{tab:cifar_chain_cons_0.25}
\begin{center}
\begin{small}
\begin{sc}
\begin{tabular}{@{}lccccc|c@{}}
\toprule
\textbf{Architecture} & \textbf{Chain Length} & \textbf{0.00} & \textbf{0.25} & \textbf{0.50} & \textbf{0.75} & \textbf{ACR}\\
\midrule
\resnext~\cite{xie2017aggregated} & 1 & 75.72 & 68.46 & 58.85 & 47.30 & 0.557 \\
\dla~\cite{yu2018deep} & 2 & 76.36 & 69.20 & 59.35 & 48.46 & 0.565 \\
\regnet~\cite{radosavovic2020designing} & 3 & 76.07 & 68.54 & 58.66 & 47.58 & 0.559 \\
\vit~\cite{dosovitskiy2020image} & 4 & 74.42 & 66.18 & 55.94 & 44.88 & 0.535 \\
\bottomrule
\end{tabular}
\end{sc}
\end{small}
\end{center}
\end{table*}


\begin{table*}[h]
\caption{\imagenet result using \sys and \smoothmix on an additional student-teacher pair. We report the ACR of the \iresnetl teacher and its \iregnet student. For reference, we also report robustness of a \iregnet network trained independently using \smoothmix. The noise level $\sigma$ is set to $0.5$. \sys remains effective on \imagenet even with a different student-teacher pair.}
\label{tab:imagenet_extra}
\begin{center}
\begin{small}
\begin{sc}
\begin{tabular}{@{}lcclc@{}}
\toprule
\multicolumn{2}{c}{\textbf{Teacher (\iresnetl)}} & & \multicolumn{2}{c}{\textbf{Student (\iregnet)}} \\ \cmidrule(lr){1-2} \cmidrule(lr){4-5}
\textbf{Training Method} & \textbf{ACR} & & \textbf{Training Method} & \textbf{ACR} \\
\midrule
\smoothmix~\cite{jeong2021smoothmix} & 0.799 & $\rightarrow$ & \sys & 0.788 \\ \cmidrule(lr){1-5}
Student trained directly & & & \smoothmix~\cite{jeong2021smoothmix} & 0.877 \\
\bottomrule
\end{tabular}
\end{sc}
\end{small}
\end{center}
\end{table*}

\begin{table*}[h]
\caption{Certified robustness and total time of a \cresnetl classifier trained on \cifar using Gaussian data augmentation. The noise level $\sigma$ is set to $0.25$.}
\vspace{1em}
\label{tab:cifar_gda}
\begin{center}
\begin{small}
\begin{sc}
\begin{tabular}{@{}lcccc|c|c}
\toprule
\textbf{Network} & \textbf{0.00} & \textbf{0.25} & \textbf{0.50} & \textbf{0.75} & \textbf{ACR} & \textbf{Total Time (h)}\\
\midrule
\cresnetl & 0.486 & 81.41 & 67.75 & 49.67 & 32.37 & 4.80 \\
\resnext & 79.71 & 66.06 & 48.67 & 31.09 & 0.474 & 4.55 \\
\dla & 81.30 & 69.53 & 54.48 & 37.81 & 0.521 & 3.08 \\
\regnet & 80.53 & 67.05 & 50.32 & 32.72 & 0.487 & 3.05 \\
\vit & 0.211 & 48.77 & 32.70 & 18.78 & 8.98 & 4.78 \\
\bottomrule
\end{tabular}
\end{sc}
\end{small}
\end{center}
\vspace{-1em}
\end{table*}

\section{Training Details} \label{app:training_details}
In this section, we provide all these details required to reproduce the results presented in the paper. We begin by reporting the hyperparameters in Appendix~\ref{app:hyperparams}, followed by code links and other necessary instructions in Appendix~\ref{app:code}.

\subsection{Hyperparamters}  \label{app:hyperparams}
First, we provide details regarding training hyperparameters for our experiments in Table~\ref{tab:train_hyperparams}. For all training, we perform regularization using a weight decay factor of $1e-4$. Also for all training, learning rate is decayed by a factor of $0.1$ at two pre-determined epochs (see column `LR Decay' in Table~\ref{tab:train_hyperparams}). Next, we report method-specific hyperparameters in Table~\ref{tab:method-hyperparams}. Note that \sys does \textbf{NOT} introduce any new hyperparameters.

\subsection{Reproducing Results From This Paper}  \label{app:code}
For reproducing results using \smoothadv~\footnote{\url{https://github.com/Hadisalman/smoothing-adversarial}}, \macer~\footnote{\url{https://github.com/RuntianZ/macer}}, \smoothmix~\footnote{\url{https://github.com/jh-jeong/smoothmix}}, and Consistency~\footnote{\url{https://github.com/jh-jeong/smoothing-consistency}}, we follow instructions provided by the authors and use their respective codes.
For \sys, all necessary instructions and code required to reproduce the results are available at \url{https://github.com/Ethos-lab/crt-neurips22}.

\begin{table}[h]
\caption{Training hyperparameters used to train classifiers using different methods. For prior works, we closely follow the hyperparameters reported by them. For \sys, we tune hyperparameters to train till convergence.}
\vspace{-0.5em}
\label{tab:train_hyperparams}
\begin{center}
\begin{threeparttable}
\begin{small}
\begin{sc}
\begin{tabular}{@{}lcccc@{}}
\toprule
\textbf{Method} & \textbf{Epochs} & \textbf{Batch Size} & \textbf{Initial LR} & \textbf{LR Decay} \\
\midrule
\multicolumn{5}{c}{\cifar} \\
\midrule
\smoothadv~\cite{salman2019provably} & 150 & 256 & 0.1 & 50, 100 \\
\macer~\cite{zhai2020macer} & 440 & 64 & 0.01 & 200, 400 \\
\smoothmix~\cite{jeong2021smoothmix} & 150 & 256 & 0.1\tnote{*} & 50, 100 \\
Gaussian Augmentation~\cite{cohen2019certified} & 200 & 128 & 0.1 & 100, 150 \\
\sys & 200 & 128 & 0.1 & 100, 150 \\
\midrule
\multicolumn{5}{c}{\imagenet} \\
\midrule
\smoothadv~\cite{salman2019provably} & 90 & 400 & 0.1 & 30,60 \\
\macer~\cite{zhai2020macer} & 120 & 256 & 0.1 & 30,60,90 \\
\smoothmix~\cite{jeong2021smoothmix} & 90 & 400 & 0.1 & 30,60 \\
\sys & 90 & 400 & 0.1 & 30,60 \\
\bottomrule
\end{tabular}
\end{sc}
\end{small}
\begin{tablenotes}
\item[*] \footnotesize{\textit{For \smoothmix training of \vit, we use initial LR of $0.01$ as otherwise training doesn't converge.}}
\end{tablenotes}
\end{threeparttable}
\end{center}
\vspace{-1em}
\end{table}

\begin{table*}[t]
\caption{Method-specific hyperparameters used in our experiments on \cifar and \imagenet.}
\label{tab:method-hyperparams}
\vskip 0.15in
\begin{center}
\begin{small}
\begin{sc}
\begin{tabular}{@{}ccc@{}}
\toprule
$\sigma$ & \textbf{Method} & \textbf{Hyperparameter Details} \\
\midrule
\multicolumn{3}{c}{\cifar} \\
\midrule
\multirow{4}{*}{0.25} & \smoothadv~\cite{salman2019provably} & 8-samples, 10-step PGD attack with $\epsilon=1.0$ \\
 & \macer~\cite{zhai2020macer} & $k=16$, $\lambda=12.0$, $\beta=16.0$, $\gamma=8.0$ \\
 & \smoothmix~\cite{jeong2021smoothmix} & $T=4$, $m=2$, $\eta=5.0$, $\alpha=0.5$ \\
 & Consistency~\cite{jeong2020consistency} & $\lambda=20$, $m=2$, $\eta=0.5$ \\
\cmidrule(lr){1-3}
0.5 & \smoothmix~\cite{jeong2021smoothmix} & $T=4$, $m=2$, $\eta=5.0$, $\alpha=1.0$ \\
\cmidrule(lr){1-3}
1.0 & \smoothmix~\cite{jeong2021smoothmix} & $T=4$, $m=2$, $\eta=5.0$, $\alpha=2.0$ \\
\midrule
\multicolumn{3}{c}{\imagenet} \\
\midrule
 & \smoothadv~\cite{salman2019provably} & 1-sample, 1-step PGD attack with $\epsilon=1.0$ \\
0.5 & \macer~\cite{zhai2020macer} & $k=2$, $\lambda=3.0$, $\beta=16.0$, $\gamma=8.0$ \\
 & \smoothmix~\cite{jeong2021smoothmix} & $T=1$, $m=1$, $\eta=1.0$, $\alpha=8.0$ \\
\bottomrule
\end{tabular}
\end{sc}
\end{small}
\end{center}
\end{table*}

\end{document}